\documentclass[letterpaper, 10 pt, conference]{ieeeconf}

\IEEEoverridecommandlockouts
\makeatletter
\let\NAT@parse\undefined
\makeatother
\usepackage[colorlinks, citecolor=green]{hyperref}
\usepackage{booktabs}
\usepackage{multirow}
\usepackage{siunitx}
\usepackage{graphicx}
\usepackage{fontawesome}
\usepackage{utfsym}
\usepackage{stfloats}
\usepackage{makecell}

\title
{
    \LARGE
    \bf
    CoSEC: A Coaxial Stereo Event Camera Dataset \\
    for Autonomous Driving
}

\author
{
    Shihan Peng$^{\dagger}$, Hanyu Zhou$^{\dagger}$, Hao Dong, Zhiwei Shi, Haoyue Liu, Yuxing Duan, Yi Chang* and Luxin Yan
    \thanks{*Corresponding author.}
    \thanks{$^{\dagger}$These authors contributed equally to this work.}
    \thanks
    {
        Shihan Peng, Hanyu Zhou, Hao Dong, Zhiwei Shi, Haoyue Liu, Yuxing Duan, Yi Chang and Luxin Yan are with National Key Lab of Multispectral Information Intelligent Processing Technology, School of Artificial Intelligence and Automation, Huazhong University of Science and Technology, Wuhan, China. Email: {\tt\small \{pengshihan, hyzhou, donghao0205, shizhiwei, liuhy, duanyuxing, yichang, yanluxin\}@hust.edu.cn}
    }
}

\begin{document}
    \maketitle
    \thispagestyle{empty} 
    \pagestyle{empty}

    
    \begin{abstract}
        Conventional frame camera is the mainstream sensor of the autonomous driving scene perception, while it is limited in adverse conditions, such as low light. Event camera with high dynamic range has been applied in assisting frame camera for the multimodal fusion, which relies heavily on the pixel-level spatial alignment between various modalities. Typically, existing multimodal datasets mainly place event and frame cameras in parallel and directly align them spatially via warping operation. However, this parallel strategy is less effective for multimodal fusion, since the large disparity exacerbates spatial misalignment due to the large event-frame baseline. We argue that baseline minimization can reduce alignment error between event and frame cameras. In this work, we introduce hybrid coaxial event-frame devices to build the multimodal system, and propose a coaxial stereo event camera (CoSEC) dataset for autonomous driving. As for the multimodal system, we first utilize the microcontroller to achieve time synchronization, and then spatially calibrate different sensors, where we perform intra- and inter-calibration of stereo coaxial devices. As for the multimodal dataset, we filter LiDAR point clouds to generate depth and optical flow labels using reference depth, which is further improved by fusing aligned event and frame data in nighttime conditions. With the help of the coaxial device, the proposed dataset can promote the all-day pixel-level multimodal fusion. Moreover, we also conduct experiments to demonstrate that the proposed dataset can improve the performance and generalization of the multimodal fusion. 
    \end{abstract}
    
    
    \begin{figure}
        \centering
        \includegraphics[width=1.0\linewidth]{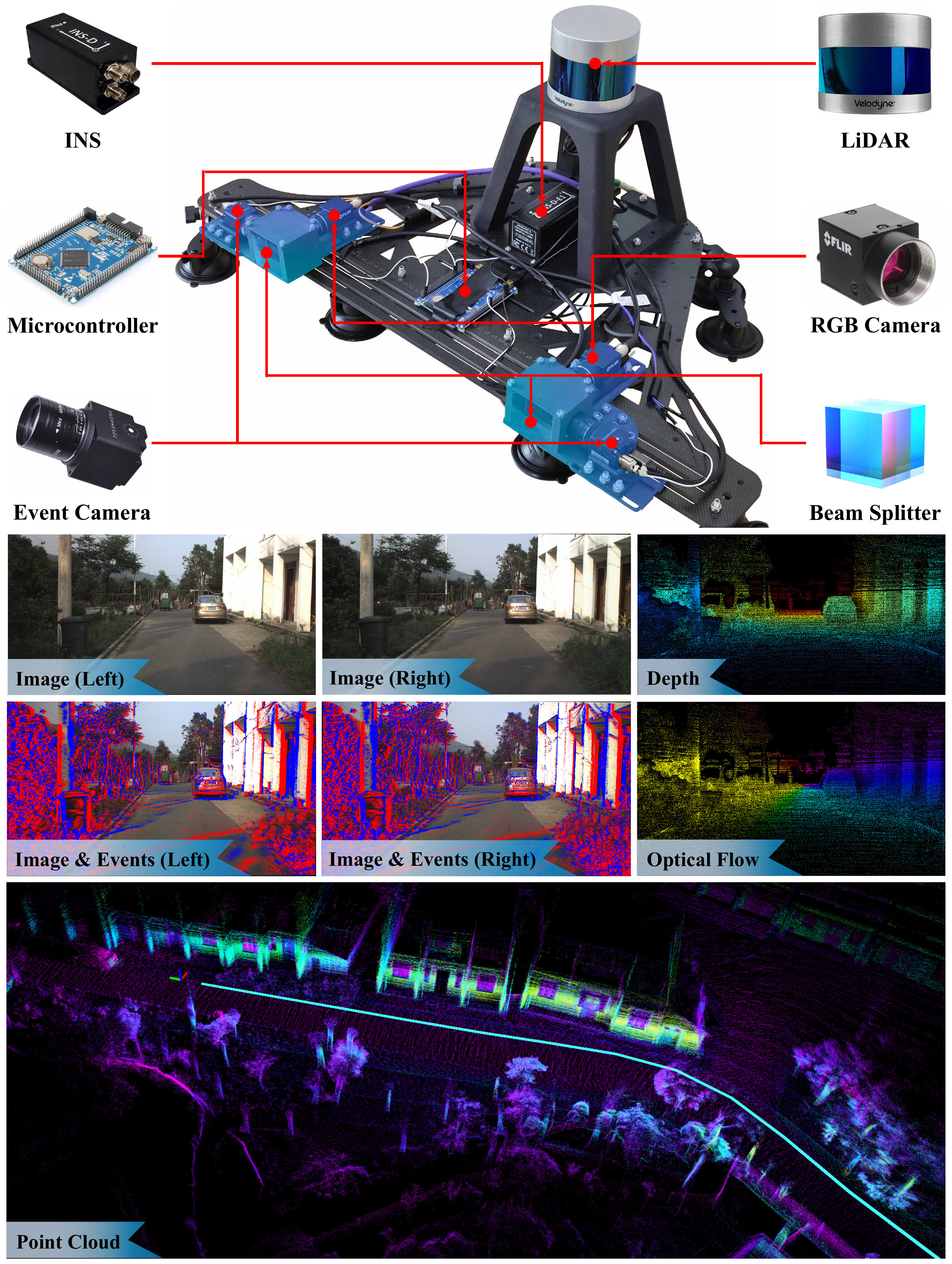}
        \caption
        {
            Illustration of the multimodal system and dataset. As for the multimodal system, we introduce the beam splitter to design the coaxial event-frame device for the pixel-level spatial alignment, and then build the coaxial stereo multimodal imaging system with the LiDAR and INS. As for the multimodal dataset, we fuse the aligned event-image data and the LiDAR point cloud to generate the ground truth depth and optical flow. In this work, we utilize the multimodal system to collect the coaxial stereo event camera dataset for autonomous driving. 
        }
        \label{Fig:Multimodal_System}
    \end{figure}
    
    \section{INTRODUCTION}
        Multimodal fusion \cite{prakash2021multi} is the key to autonomous driving scene perception, aiming to fuse the complementary knowledge between frame, LiDAR, and inertial measurement unit (IMU) to perceive the 3D dynamic scene. Conventional frame camera performs well in ideal imaging conditions, while suffering degradation in adverse imaging conditions, such as low light \cite{liang2023coherent} and fast motion \cite{liu2024seeing}. The main reason is that, the frame camera is a fixed-exposure imaging mechanism, which is limited by its low dynamic range and low frame rate. In contrast, the event camera \cite{gallego2020event} with the advantages of high dynamic range and high temporal resolution has been applied in assisting frame camera for multimodal fusion, which relies heavily on the pixel-level spatial alignment between various modal data. In this work, our purpose is to propose a spatiotemporal-aligned multimodal dataset for autonomous driving in Fig. \ref{Fig:Multimodal_System}.
        
        Existing event-based datasets for autonomous driving are mainly divided into three categories: event-only unimodal, event-frame multimodal and stereo event-frame multimodal datasets. As shown in Table \ref{Tab:Compdata}, event-only unimodal datasets \cite{sironi2018hats, de2020large, perot2020learning, cheng2019det} only focus on relatively simple vision tasks, while being difficult to model ego-motion and 3D dynamic scene. Event-frame multimodal datasets \cite{binas2017ddd17, hu2020ddd20, fischer2020event} introduce IMU to model the ego-motion of vehicles, but still fail for 3D scene understanding. Stereo event-frame multimodal datasets \cite{zhu2018multivehicle, Gehrig21ral, wei2024fusionportablev2, chaney2023m3ed} further introduce LiDAR and stereo cameras to improve 3D dynamic scene perception. These multimodal datasets mainly place conventional frame camera and event camera in parallel, and spatially align them by warping operation \cite{gehrig2024low}. However, this parallel strategy cannot guarantee the accurate spatial alignment between various modalities, which is less effective for multimodal fusion. Therefore, how to ensure the \emph{event-frame pixel-level alignment} is crucial for multimodal fusion.
        
        As for the pixel-level spatial alignment, we explore the impact of the baseline between event and frame cameras on it in Fig. \ref{Fig:Devices_Compare}, illustrating that the large baseline deteriorates the event-frame spatial alignment. The main reason is that, the large baseline between event and frame cameras brings in the large disparity, increasing the pixel shift of the corresponding points between the two cameras, thus restricting the spatial alignment. Therefore, we suggest that \emph{baseline minimization reduces the event-frame pixel-level spatial alignment error}.
        
        \begin{table*}
            \vspace{5pt}
            \caption
            {
                Comparison of different event camera datasets for autonomous driving. Event-only unimodal datasets provide single event data for relatively simple vision tasks. Event-frame multimodal datasets introduce frame cameras and INS sensors for motion perception. Stereo event-frame multimodal datasets further introduce LiDAR for 3D scene motion perception. In this work, we further design a coaxial event-frame strategy to improve the pixel-level alignment of the multimodal dataset.
            }
            \label{Tab:Compdata}
            \centering
            \setlength\tabcolsep{3.5pt}
            \renewcommand{\arraystretch}{1.15}
            \begin{tabular}{clcccccccc}
                \toprule
                    \multicolumn{1}{c}{\multirow{2.5} * {\textbf{Type}}} & 
                    \multicolumn{1}{c}{\multirow{2.5} * {\textbf{Dataset}}} & 
                    \multirow{2.5} * {\textbf{LiDAR}} & \multirow{2.5} * {\textbf{IMU}} & \multirow{2.5} * {\textbf{GNSS}} & 
                    \multicolumn{2}{c}{\textbf{Frame Camera}} & \multicolumn{2}{c}{\textbf{Event Camera}} & 
                    \multirow{2.5} * {\textbf{Groundtruth}} \\
                    \cmidrule(lr){6-7} \cmidrule(lr){8-9} & & & & & 
                    \textbf{Resolution} & \textbf{Color} & \textbf{Resolution} & \textbf{Aligned with Frame} & \\
                \midrule
                    \multirow{4} * {\makecell{\textbf{Event-only} \\ \textbf{Unimodal}}}
                    & N-CARS \cite{sironi2018hats} & 
                        \usym{2613} & \usym{2613} & \usym{2613} & 
                        \multicolumn{2}{c}{\usym{2613}} & ~304 $\times$ 240 & 
                        \usym{2613} & Classification \\
                    & ADD \cite{de2020large} & 
                        \usym{2613} & \usym{2613} & \usym{2613} & 
                        \multicolumn{2}{c}{\usym{2613}} & ~304 $\times$ 240 & 
                        \usym{2613} & Detection \\
                    & 1MP Detection \cite{perot2020learning} &
                        \usym{2613} & \usym{2613} & \usym{2613} & 
                        \multicolumn{2}{c}{\usym{2613}} & 1280 $\times$ 720 & 
                        \usym{2613} & Detection \\
                    & DET \cite{cheng2019det} & 
                        \usym{2613} & \usym{2613} & \usym{2613} & 
                        \multicolumn{2}{c}{\usym{2613}} & 1280 $\times$ 800 & 
                        \usym{2613} & Lane Extraction \\
                \midrule
                    \multirow{4} * {\makecell{\textbf{Event-Frame} \\ \textbf{Multimodal}}}
                    & DDD17 \cite{binas2017ddd17} & 
                        \usym{2613} & \faCheck & \faCheck & 
                        ~346 $\times$ 260 & \usym{2613} & ~346 $\times$ 260 & 
                        \faCheck & Vehicle Control \\
                    & DDD20 \cite{hu2020ddd20} & 
                        \usym{2613} & \faCheck & \faCheck & 
                        ~346 $\times$ 260 & \usym{2613} & ~346 $\times$ 260 & 
                        \faCheck & Vehicle Control \\
                    & \multirow{2} * {Brisbane-Event-VPR \cite{fischer2020event}} & 
                        \multirow{2} * {\usym{2613}} & \multirow{2} * {\faCheck} & \multirow{2} * {\faCheck} & 
                        ~346 $\times$ 260 & \faCheck & \multirow{2} * {~346 $\times$ 260} & 
                        \faCheck & \multirow{2} * {Place Recognition} \\ [-1ex]
                        & & & & & ~1920 $\times$ 1080 & \faCheck & & \usym{2613} & \\
                \midrule
                    \multirow{7} * {\makecell{\textbf{Stereo} \\ \textbf{Event-Frame} \\ \textbf{Multimodal}}}
                    & \multirow{2} * {MVSEC \cite{zhu2018multivehicle}} & 
                        \multirow{2} * {\faCheck} & \multirow{2} * {\faCheck} & \multirow{2} * {\faCheck} & 
                        ~346 $\times$ 260 & \multirow{2} *{\usym{2613}} & \multirow{2} * {~346 $\times$ 260} & 
                        \faCheck & \multirow{2} * {Depth, Optical Flow} \\ [-1ex]
                        & & & & & ~752 $\times$ 480 & & & \usym{2613} & \\
                    & \multirow{2} * {FusionPortableV2 \cite{wei2024fusionportablev2}} & 
                        \multirow{2} * {\faCheck} & \multirow{2} * {\faCheck} & \multirow{2} * {\faCheck} & 
                        ~346 $\times$ 260 & \multirow{2} * {\faCheck} & \multirow{2} * {~346 $\times$ 260} & 
                        \faCheck & \multirow{2} * {SLAM} \\ [-1ex]
                        & & & & & 1024 $\times$ 768 & & & \usym{2613} & \\
                    & DSEC \cite{Gehrig21ral} & 
                        \faCheck & \faCheck & \faCheck & 
                        ~1440 $\times$ 1080 & \faCheck & ~640 $\times$ 480 & 
                        \usym{2613} & Depth, Optical Flow \\ 
                    & M3ED \cite{chaney2023m3ed} & 
                        \faCheck & \faCheck & \faCheck & 
                        1280 $\times$ 800 & \faCheck & 1280 $\times$ 720 & 
                        \usym{2613} & Depth, Optical Flow \\ 
                \cmidrule(lr){2-10}
                    & CoSEC (Ours) & 
                        \faCheck & \faCheck & \faCheck & 
                        ~2048 $\times$ 1536 & \faCheck & 1280 $\times$ 720 & 
                        \faCheck~(1200 $\times$ 624) & Depth, Optical Flow \\
                \bottomrule
            \end{tabular}
        \end{table*}
        
        In this work, we introduce a hybrid coaxial event-frame device to build the multimodal system in Fig. \ref{Fig:Multimodal_System}, and propose a coaxial stereo event camera (CoSEC) dataset for autonomous driving, including frame, event, LiDAR, IMU and RTK GNSS. As for the system, we realize time synchronization and spatial calibration between various modalities. During time synchronization, we utilize the microcontroller and GNSS to jointly calibrate the timestamp. During spatial calibration, we first perform intra-calibration within single coaxial device and inter-calibration between stereo coaxial devices, and then further calibrate other sensors. As for the dataset, we collect multimodal data covering all-day scenes. To generate ground truth depth and optical flow, we use SLAM algorithm \cite{xu2021fast} to fuse LiDAR point clouds, and filter them with reference depth estimated by a foundation model \cite{yin2023metric3d, hu2024metric3d}. Moreover, we additionally fuse the aligned event-frame data to enhance the nighttime imaging performance, thus improving the accuracy of the ground truth. With the help of the coaxial device, the proposed dataset can promote all-day pixel-level multimodal fusion. Overall, our main contributions
        can be summarized as follows:
        \begin{itemize}
            \item We introduce coaxial event-frame devices to build a multimodal system, which realizes the spatiotemporal alignment between various modalities, thus promoting multimodal fusion perception.
            
            \item We propose a coaxial stereo event camera dataset covering all-day scenes, which fuses the aligned event and frame data to improve the accuracy of depth and optical flow labels in nighttime conditions.
            
            \item We conduct experiments to demonstrate that pixel-level aligned multimodal data can improve the performance and generalization of multimodal fusion.
        \end{itemize}
    
    
    \section{RELATED WORK}
        Event camera \cite{gallego2020event} has garnered increasing attention due to its advantages of high dynamic range and high temporal resolution. It has been applied in different autonomous driving datasets in Table \ref{Tab:Compdata}, including event-only unimodal, event-frame multimodal and stereo event-frame multimodal datasets. In this section, we will describe the development of these event-based datasets in detail.
        
        \subsection{Event-only Unimodal Datasets}
            Event-only unimodal datasets \cite{sironi2018hats, de2020large, perot2020learning, cheng2019det} just utilize the event camera to perform relatively simple vision tasks in autonomous driving. For example, N-CARS \cite{sironi2018hats} and ADD \cite{de2020large} datasets took Gen1 event camera with 304$\times$240 resolution to classify and detect objects. 1MP Detection dataset \cite{perot2020learning} further introduced EVK4 event camera with a higher resolution of 1280$\times$720 for object detection. DET dataset \cite{cheng2019det} focused on lane extraction with an event camera. However, these unimodal datasets are not sufficiently suitable for the challenging task of motion perception, which is the key to understanding the dynamic scene for autonomous driving. Therefore, our dataset focuses on motion perception in driving scenarios, such as optical flow. 

        \begin{figure}
            \vspace{5pt}
            \centering
            \includegraphics[width=1.0\linewidth]{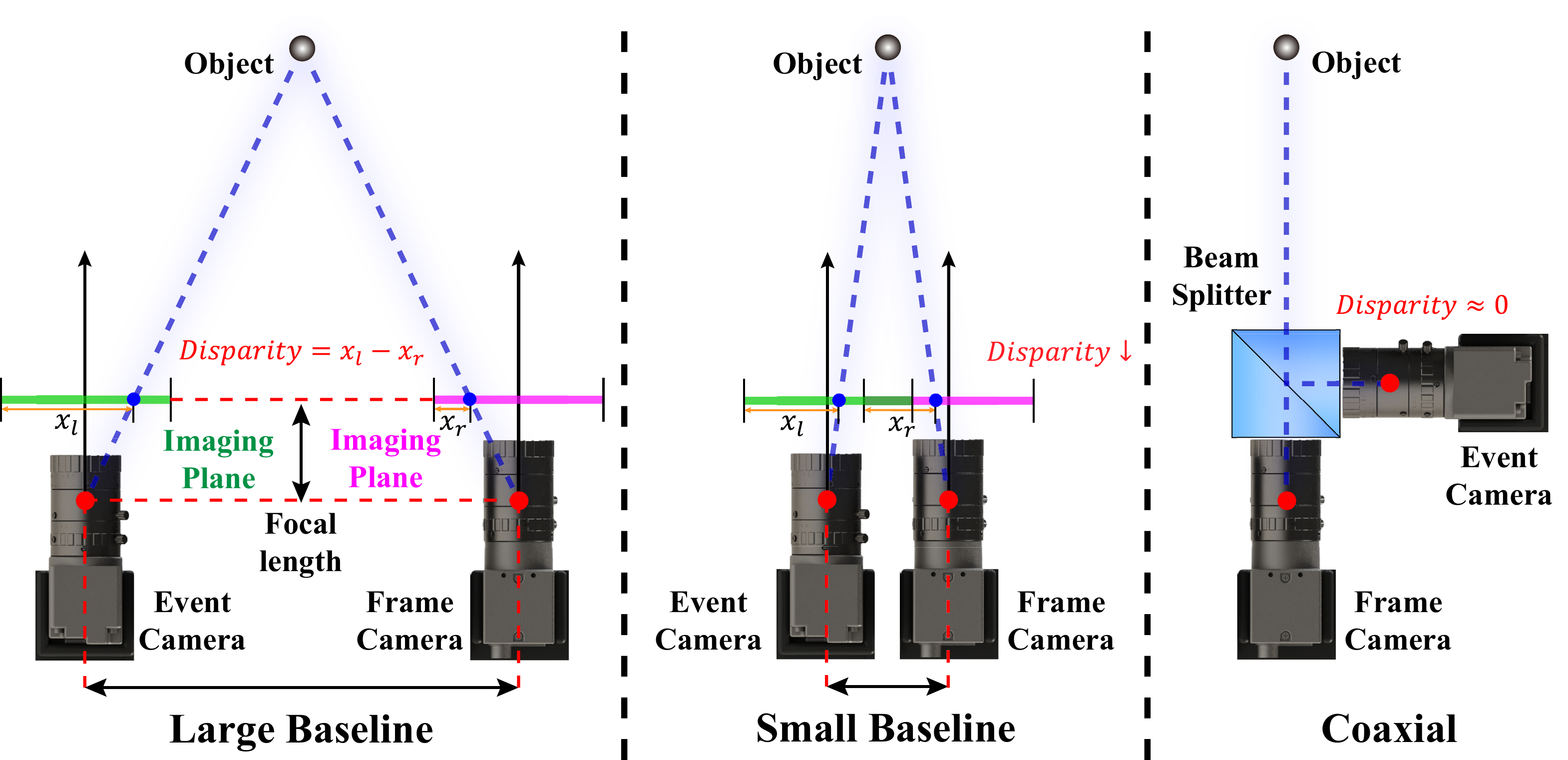}
            \caption
            {
                Difference between different event-frame placement strategies. Left: The large baseline leads to the large disparity, resulting in the misalignment between event and frame camera. Middle: The small baseline brings in a small but non-negligible disparity. Right: Baseline minimization can reduce the disparity. Therefore, we introduce a coaxial strategy to relieve the spatial alignment error between event and frame camera.
            }
            \label{Fig:Devices_Compare}
        \end{figure}
        
        \subsection{Event-Frame Multimodal Datasets}           
            Event-frame multimodal datasets \cite{binas2017ddd17, hu2020ddd20, fischer2020event} integrate frame camera and sensors of vehicle control (\emph{e.g.}, IMU ) to fuse cross-modal complementary knowledge for motion perception in autonomous driving. For example, DDD17 \cite{binas2017ddd17} and DDD20 \cite{hu2020ddd20} datasets fused appearance information of road direction from event and frame data to predict all-day vehicle steering motion. Brisbane-Event-VPR dataset \cite{fischer2020event} maximizes the similarity between features extracted from event and frame data for dynamic scene understanding, such as place recognition for autonomous driving in challenging lighting conditions. These datasets demonstrate the advantages of multimodal fusion for robust motion perception under various illumination. However, these datasets are inadequate for 3D scene perception in autonomous driving due to the absence of distance sensors, such as LiDAR. In this work, we further focus on the multimodal dataset for motion perception of the 3D dynamic scene in autonomous driving.
             
        \subsection{Stereo Event-Frame Multimodal Datasets}
            Stereo event-frame multimodal datasets \cite{zhu2018multivehicle, Gehrig21ral, wei2024fusionportablev2, chaney2023m3ed} improve 3D dynamic scene motion perception using the distance sensors. For example, the MVSEC \cite{zhu2018multivehicle} and FusionPortableV2 \cite{wei2024fusionportablev2} datasets introduced the LiDAR and stereo DAVIS event camera \cite{brandli2014240, li2015design, taverni2018front} to fuse the multimodal knowledge for 3D perception, such as depth estimation and SLAM. The DSEC \cite{Gehrig21ral} and M3ED \cite{chaney2023m3ed} datasets further utilized the Prophesee event camera with a higher resolution for more fine-grained stereo matching in driving scenarios. Nevertheless, these datasets commonly place event and frame cameras in parallel, with a small but non-negligible baseline between them as shown in Fig. \ref{Fig:Devices_Compare}. This parallel strategy introduces a disparity that cannot be ignored, making it impossible to ensure accurate spatial alignment between event and frame cameras. Therefore, our CoSEC dataset introduces the beam splitter to minimize the baseline between event and frame cameras, building a coaxial event-frame device to collect spatiotemporal-aligned event-frame data with a resolution of 1200$\times$624.

        \begin{figure}
            \vspace{5pt}
            \centering
            \includegraphics[width=1.0\linewidth]{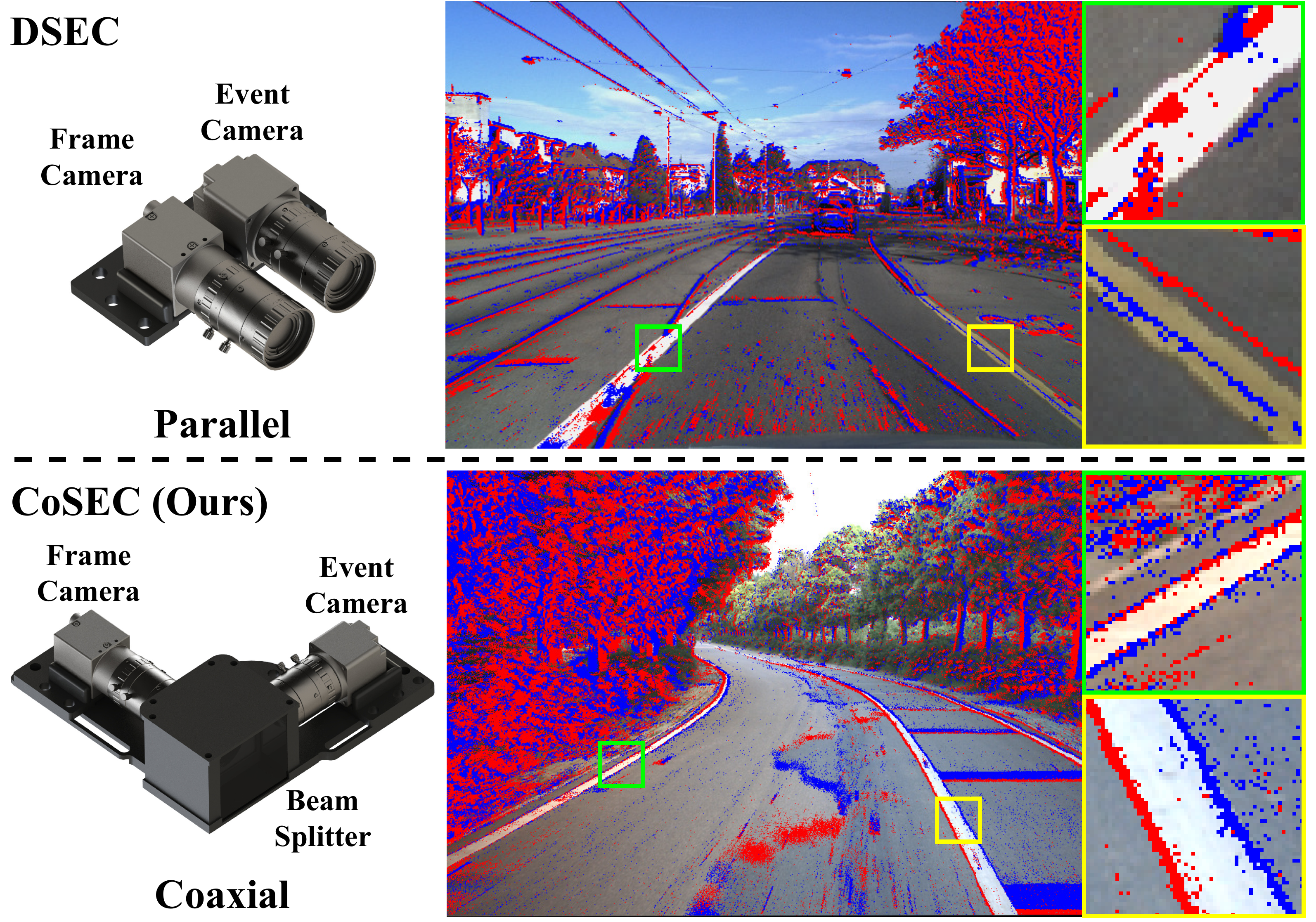}
            \caption
            {
                Comparison between parallel and coaxial strategies. Parallel placement strategy brings in spatial alignment error between the event and frame data in local regions. In contrast, we introduce a coaxial strategy to improve the pixel-level spatial alignment between the event and frame data.
            }
            \label{Fig:Data_Compare}
        \end{figure}

    
    \section{MOTIVATION OF COAXIAL DEVICE}
        The prerequisite for multimodal fusion is the pixel-level spatial alignment between different modalities. Existing multimodal datasets mainly place frame and event cameras in parallel, and directly spatially align the two cameras via warping operation \cite{gehrig2024low}. However, this parallel strategy cannot guarantee the accurate spatial alignment between various modalities, which is less effective for multimodal fusion. As shown in Fig. \ref{Fig:Devices_Compare}, we explore the impact of different baselines on spatial alignment between event and frame cameras. We can observe that, the large baseline brings in the large disparity, which is indeed beneficial for scene measurement of stereo cameras, but restricts the cross-modal spatial alignment due to the increased pixel shift of the corresponding points. Conversely, the smaller the baseline, the smaller the disparity and the higher the degree of pixel-level spatial alignment between event and frame cameras.
        
        Motivated by this, we minimize the event-frame baseline using the coaxial strategy, which uses a beam splitter to ensure that the event and frame cameras share the same optical axis for physically reducing the spatial alignment error. Moreover, we further compare the alignment performance of the parallel strategy from DSEC \cite{Gehrig21ral} dataset and our coaxial strategy in Fig. \ref{Fig:Data_Compare}, where we use the warping operation similar to \cite{gehrig2024low}. We can observe that, there exists obvious pixel-level misalignment between event data and frame data aligned through the parallel strategy, while the coaxial strategy can significantly improve the accuracy of global alignment. In this work, we will introduce a coaxial event-frame device to build the multimodal system for pixel-level spatial alignment.
    
    
    \section{MULTIMODAL SYSTEM}
        Different sensors of the multimodal system have their own specific imaging characteristics, and it is difficult to directly obtain all modal data in the same scene. How to ensure the pixel-level alignment of different sensors in the temporal and spatial dimensions is the key to the multimodal system.
        
        \subsection{Sensors and Time Synchronization}
            \textbf{Sensors.} We first design the 3D structure of the multimodal system consisting of several sensors in Table \ref{Tab:Sensorslist}, including stereo coaxial devices, one LiDAR and one integrated navigation system (INS). As for the stereo coaxial devices, we first choose the Prophesee EVK4 event camera with a resolution of 1280 $\times$ 720 and the FLIR Blackfly S USB3 color frame camera with a resolution of 2048 $\times$ 1536 to construct the single coaxial device, which share the same optical axis via a beam splitter for the physical spatial alignment. Then, we build the stereo camera system with the two coaxial devices. As for the LiDAR, we choose the Velodyne VLP-32C to generate the point clouds for measuring the 3D scene distance. As for the INS, we choose the InertialLabs IMU to measure the angular velocity and acceleration of ego-motion, and the Novatel OEM7 GNSS receiver to measure the latitude and longitude position. LiDAR point clouds and INS data can be fused to generate depth and optical flow labels for autonomous driving.
         
            \begin{table}
                \vspace{5pt}
                \caption{Sensors of the multimodal system}
                \label{Tab:Sensorslist}
                \centering
                \begin{tabular}{cl}
                        \toprule
                                \textbf{Sensor}
                                    & \multicolumn{1}{c}{\textbf{Characteristics}}                    \\
                        \midrule
                                \multirow{3} * {\textbf{Prophesee EVK4-HD}}
                                    & Latency: 220 \si{\micro\second}                                 \\
                                    & Resolution: 1280 $\times$ 720                                   \\
                                    & Dynamic Range: \textgreater 120 dB                              \\  
                        \midrule
                                \multirow{3} * {\textbf{FLIR BFS-U3-32S4C}}
                                    & Chroma: Color                                                   \\
                                    & Resolution: 2048 $\times$ 1536                                  \\
                                    & Dynamic Range: 71.62 dB                                         \\  
                        \midrule
                                \multirow{5} * {\textbf{Velodyne VLP-32C}}
                                    & Channels: 32                                                    \\
                                    & Measurement Range: 200 m                                        \\
                                    & Range Accuracy: $\pm$3 cm                                       \\
                                    & Horizontal Field of View: 360$^\circ$                           \\
                                    & Vertical Field of View: 40$^\circ$ (-25$^\circ$ to +15$^\circ$) \\
                        \midrule
                                \multirow{3} * {\textbf{InertialLabs INS-D-E1}} 
                                    & 9-Axis IMU                                                      \\
                                    & Novatel OEM7 GNSS receiver                                      \\
                                    & Position Accuracy (RTK): $\pm$2 cm                              \\
                        \bottomrule
                \end{tabular}
            \end{table}
            
            \textbf{Time Synchronization.} Since all the sensors have different data acquisition frequency, it is necessary to achieve time synchronization, which ensures that all the sensors capture information from the same scene at the same time. Time synchronization is divided into two steps, including external trigger and absolute time calibration. During external trigger stage, we utilize microcontroller to generate four synchronous pulses, where two are the pulses with high frequency (\emph{e.g.}, 1M Hz) for event cameras and two are the pulses with low frequency (\emph{e.g.}, 30 Hz) for frame cameras, thus achieving time synchronization between event and frame cameras. During absolute time calibration stage, we use the microcontroller to receive the pps signals from the GNSS receiver, which is further parsed into absolute timestamps for LiDAR, IMU and cameras. In this way, all the sensors work under GNSS timestamps, achieving time synchronization.

            \begin{figure}
                \vspace{5pt}
                \centering
                \includegraphics[width=1.0\linewidth]{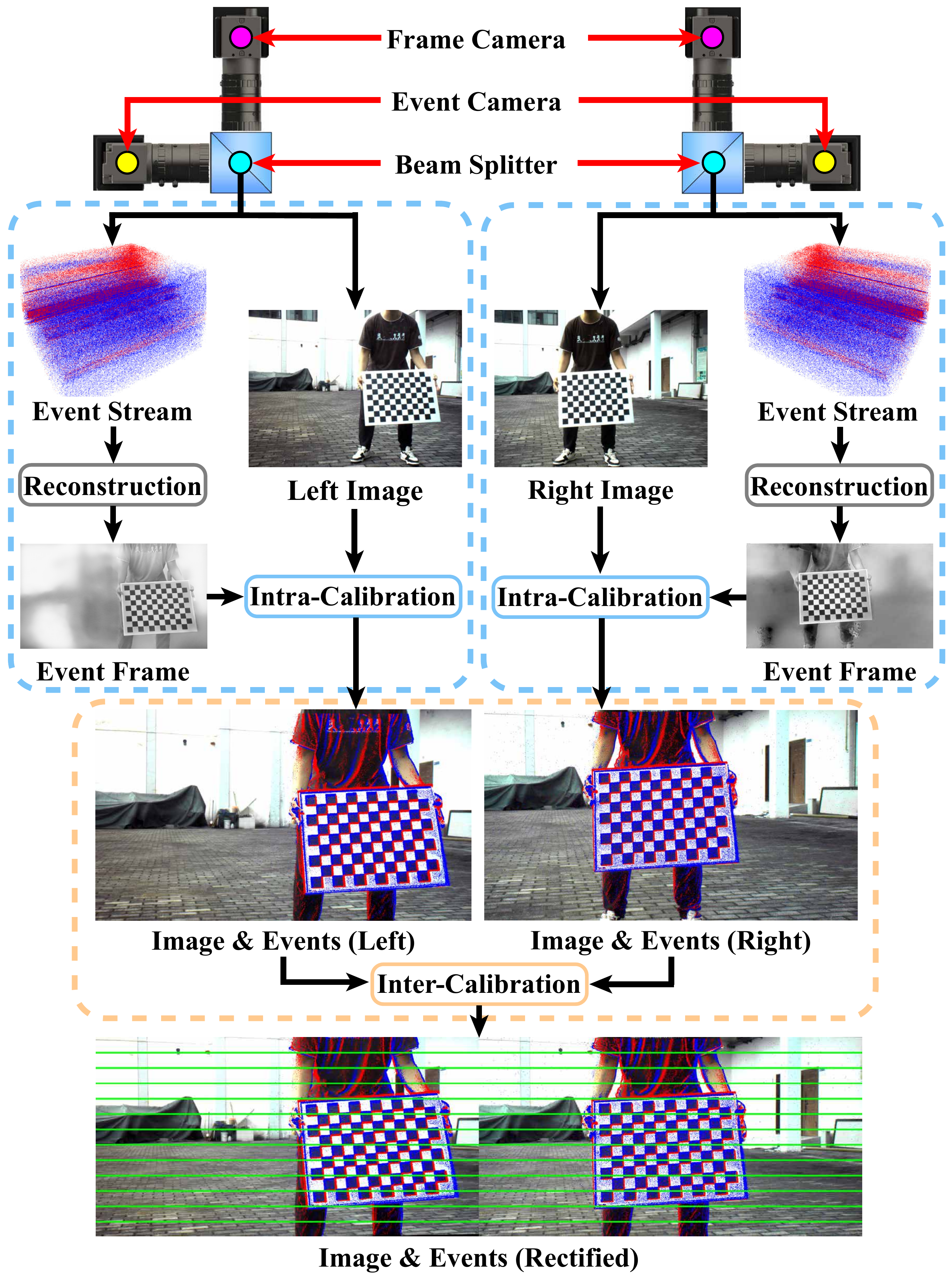}
                \caption
                {
                    Calibration of the stereo coaxial devices. We perform intra-calibration within the single coaxial device and inter-calibration between stereo coaxial devices. During intra-calibration, we first reconstruct events into event frames for standard calibration, and then align the event and image data via warping operation. During inter-calibration, we further take stereo rectification to obtain the paired rectified event-image data.
                }
                \label{Fig:Coaxial_Rectification}
            \end{figure}

    \begin{figure*}
        \vspace{5pt}
        \centering
        \includegraphics[width=1.0\linewidth]{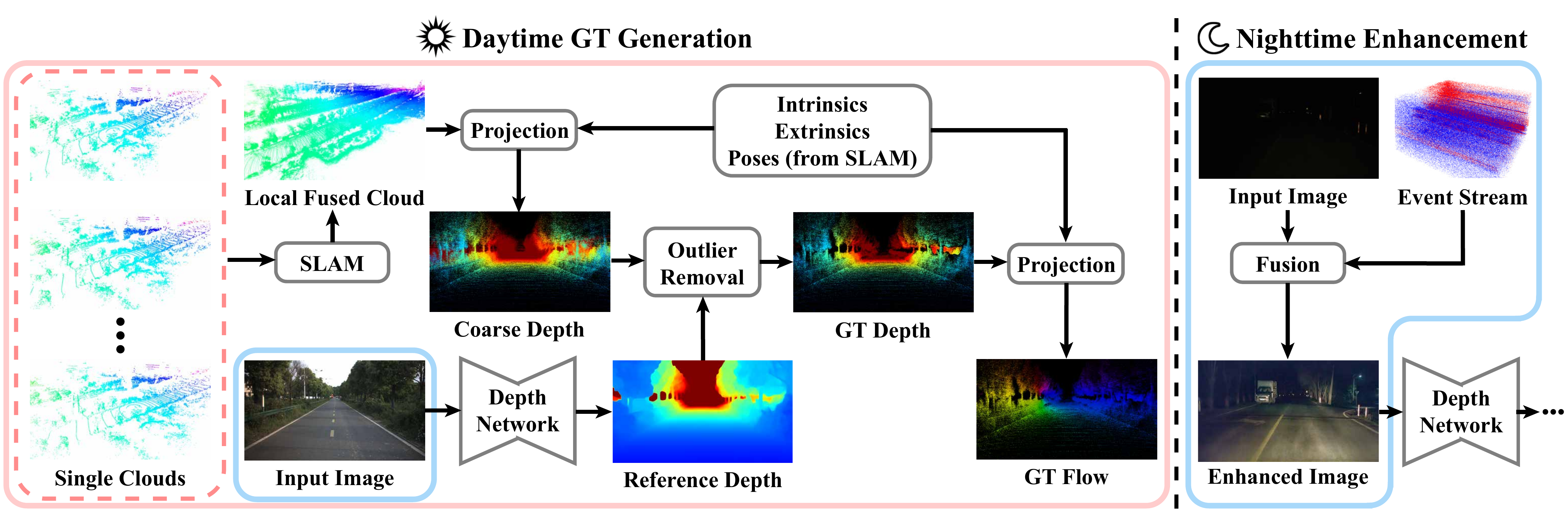}
        \caption
        {
            Pipeline of ground truth generation. We first fuse single clouds within a time window into a local cloud via SLAM, and then project the local fused cloud into the camera coordinate system for coarse depth. Next, we design an outlier removal module, which estimates reference depth from the input image to filter the coarse depth for ground truth depth and optical flow. In addition, we introduce an event-frame fusion strategy to enhance the nighttime low-light image for achieving better reference depth, thus improving the accuracy of ground truth generation.
        }
        \label{Fig:GT_Generation}
    \end{figure*}

    \begin{table}
        \caption{The overview of the CoSEC dataset.}
        \label{Tab:Sequences}
        \centering
        \begin{tabular}{ccccccccc}
            \toprule
                \textbf{Split} & \textbf{Time} & \textbf{Area} & \multicolumn{3}{c}{\textbf{Sequences}} & \multicolumn{3}{c}{\textbf{Duration (s)}} \\
            \midrule
                \multirow{10} * {\textbf{Train}} & 
                    \multirow{5} * {Day} & 
                        City & 11 & \multirow{5} * {58} & \multirow{10} * {\textbf{102}} & 320 & \multirow{5} * {1652} & \multirow{10} * {\textbf{2905}} \\
                        & & Campus & 12 & & & 338 & & \\
                        & & Park & 11 & & & 314 & & \\
                        & & Suburbs & 15 & & & 411 & & \\
                        & & Village & 9 & & & 269 & & \\
                    \cmidrule(lr){2-5} \cmidrule(lr){7-8}
                    & \multirow{5} * {Night} & 
                        City & 9 & \multirow{5} * {44} & & 253 & \multirow{5} * {1253} & \\
                        & & Campus & 10 & & & 287 & & \\
                        & & Park & 8 & & & 231 & & \\
                        & & Suburbs & 12 & & & 334 & & \\
                        & & Village & 5 & & & 148 & & \\
            \midrule
                \multirow{10} * {\textbf{Test}} & 
                    \multirow{5} * {Day} & 
                        City & 2 & \multirow{5} * {15} & \multirow{10} * {\textbf{26}} & 58 & \multirow{5} * {431}  & \multirow{10} * {\textbf{753}}  \\
                        & & Campus & 3 & & & 94 & & \\
                        & & Park & 3 & & & 84 & & \\
                        & & Suburbs & 5 & & & 138 & & \\
                        & & Village & 2 & & & 57 & & \\
                    \cmidrule(lr){2-5} \cmidrule(lr){7-8}
                    & \multirow{5} * {Night} & 
                        City & 2 & \multirow{5} * {11} & & 61 & \multirow{5} * {322} & \\
                        & & Campus & 2 & & & 66 & & \\
                        & & Park & 2 & & & 56 & & \\
                        & & Suburbs & 3 & & & 85 & & \\
                        & & Village & 2 & & & 54 & & \\
            \midrule
                \textbf{Total} & & & & & \textbf{128} & & & \textbf{3658} \\
            \bottomrule
        \end{tabular}
    \end{table}   
            
        \subsection{Spatial Calibration}            
            \textbf{Intra-Calibration within Coaxial Device.} Since event camera and frame camera have different spatial resolutions, we need to calibrate the two camera into the same imaging coordinate system within the single coaxial device, which guarantees the pixel-level spatial alignment. As shown in Fig. \ref{Fig:Coaxial_Rectification}, first, we simultaneously capture event stream and color frames for the same checkerboard pattern. Then, considering that event data cannot be directly used for the standard calibration, we reconstruct the event stream into the event frame via E2VID \cite{rebecq2019events, rebecq2019high}. Finally, we take Kalibr toolbox \cite{furgale2013unified} to calibrate the event and frame cameras using the color image and the reconstructed event frame, thus obtaining the intrinsics and extrinsics of the two cameras. Note that these calibration parameters can be used to generate aligned event-frame data by using the warping operation \cite{gehrig2024low}.

            \textbf{Inter-Calibration between Coaxial Devices.} Stereo coaxial devices consist of two coaxial event-frame devices, similar to the stereo camera system, which could be further performed inter-calibration between the two coaxial devices to obtain the paired rectified event-frame data for stereo matching. First, we calibrate the stereo frame cameras, achieving rectified stereo images with the original resolution (2048 $\times$ 1536). Next, the calibrated extrinsics between the two frame cameras are used to compute the extrinsics of the stereo coaxial devices as follows:
            \begin{equation}
                \label{ToRL}
                    T_{oRL} = T_{LL_{E}}T_{RL}T_{RR_{E}}^{-1}
            \end{equation}
            where $T_{oRL}$ is the transformation matrix from right to left coaxial device, and $T_{RL}$ is the transformation matrix from right to left frame camera. $T_{LL_{E}}$ is the transformation matrix from frame to event camera within the left coaxial device, and $T_{RR_{E}}$ is similar for the right. Note that $T_{RL}$ is obtained from the calibration of stereo frame cameras, and $T_{LL_{o}}$, $T_{RR_{o}}$ are achieved from the intra-calibration of the coaxial devices. Finally, we use $T_{oRL}$ and the camera intrinsics to rectify the aligned events and frames, obtaining paired rectified event-frame data with a resolution of 1200$\times$624.
           
            In addition, we also use the Kalibr toolbox \cite{furgale2013unified} for the calibration between the IMU and camera, where we keep the multimodal system moving towards the AprilTag. As for the calibration between the LiDAR and camera, we first obtain initial extrinsics between the LiDAR and camera from the CAD model. And then, we take the LiDAR-to-camera calibration tool from the SensorsCalibration toolbox \cite{yan2022opencalib}, to refine these initial extrinsics.
        

    \begin{figure}
        \centering
        \includegraphics[width=1.0\linewidth]{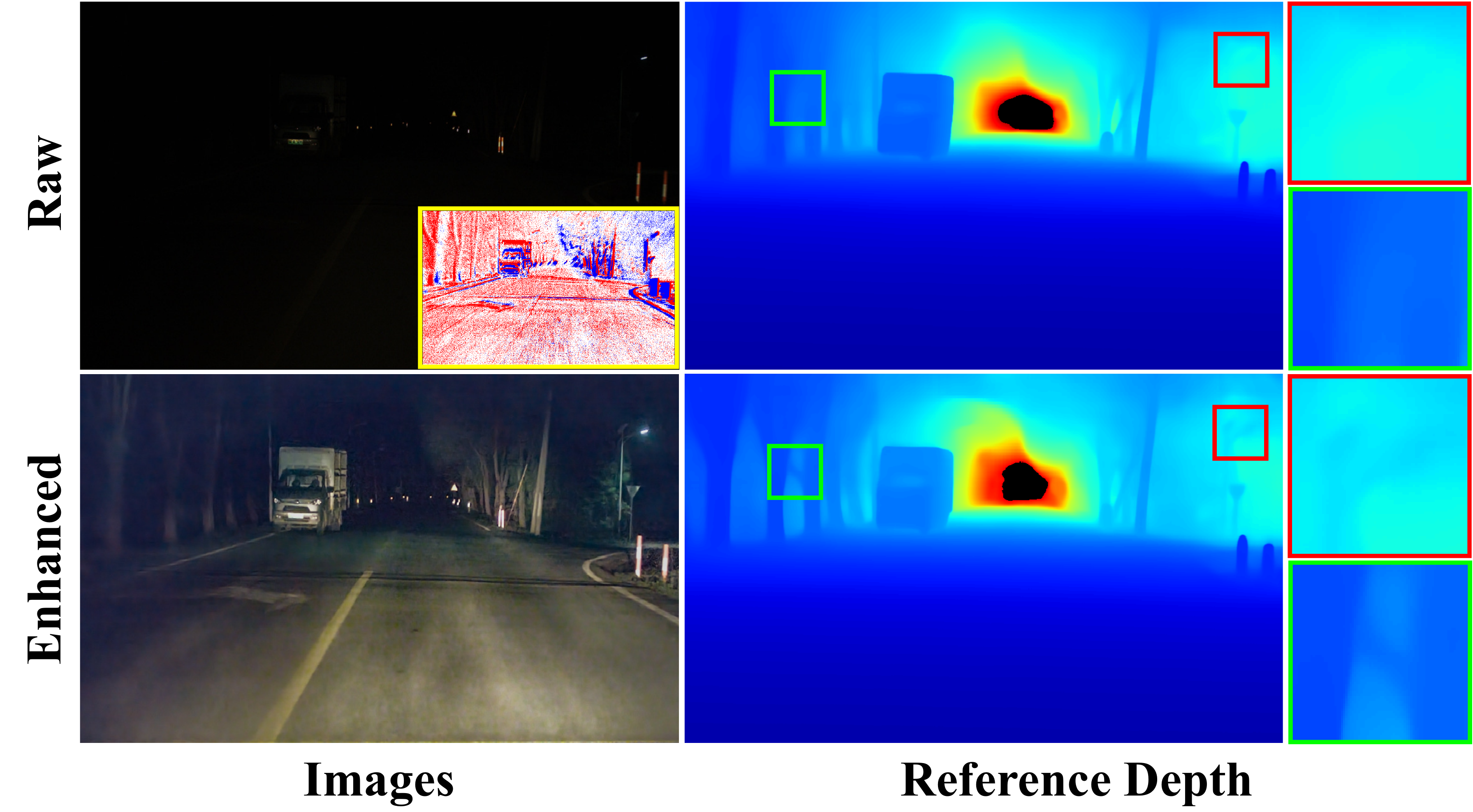}
        \caption
        {
            Effect of event-frame fusion on reference depth in the nighttime scene. Reference depth obtained from the low-light image shows local smoothness due to the lost texture. In contrast, event-frame fusion enhances the imaging quality, thus improving structure details of reference depth.
        }
        \label{Fig:Enhance_Compare}
    \end{figure} 

        \begin{figure*}
            \vspace{5pt}
            \centering
            \includegraphics[width=1.0\linewidth, height=220pt]{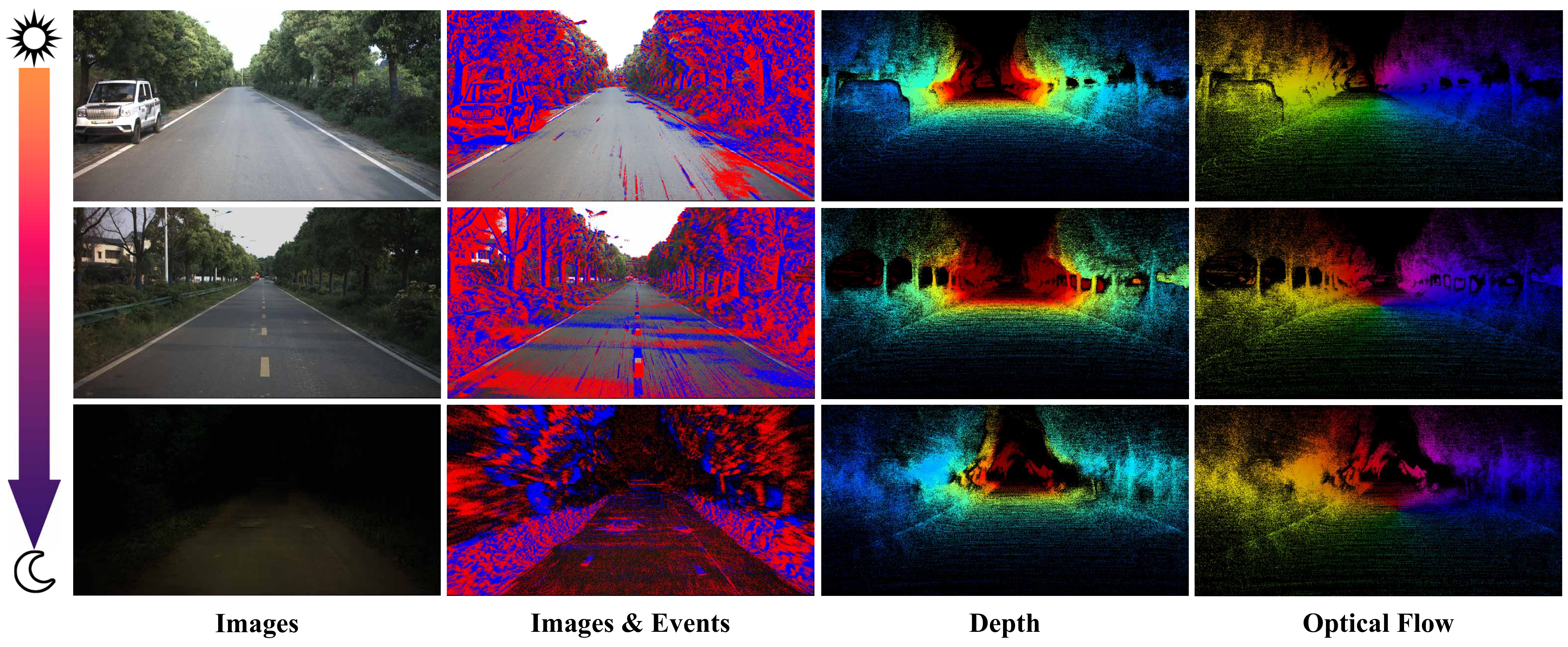}
            \caption
            {
                Ground truth examples of the proposed dataset CoSEC under various illumination conditions. CoSEC can provide the pixel-level spatiotemporal-aligned event and frame data with corresponding ground truth under ideal light and extremely low-light conditions.
            }
            \label{Fig:GT_Visual}
        \end{figure*}
        
        \section{MULTIMODAL DATASET}
        Multimodal dataset consists of input data and high-quality ground truth. The ground truths of autonomous driving datasets mainly include detection, semantics, depth, optical flow, etc. In this section, we focus on the ground truth generation techniques of the depth and optical flow labels based on physical calculation process. In addition, we also introduce the cross-modal fusion strategy for the reliability of labels in nighttime conditions.
        
        \subsection{Sequences}
            The proposed multimodal dataset CoSEC covers various scenes under all-day conditions with total 128 sequences of 3658 seconds. As shown in Table \ref{Tab:Sequences}, the sequences of CoSEC are collected in city, campus, park, suburbs and village under daytime and nighttime conditions. These sequences are split into 102 for training and 26 for testing. Moreover, each sequence provides the stereo spatiotemporal-aligned event and frame data with their corresponding ground truth depth and optical flow. The procedure for ground truth generation will be presented in the following section.
           
        \subsection{Ground Truth Generation}
            Autonomous driving scene perception usually focuses on the 3D semantic and motion information of dynamic scenes, where the former is obtained through manual annotation while the latter is obtained through data calculation. In this part, we mainly describe the ground truth depth and optical flow generation in Fig. \ref{Fig:GT_Generation}.

            \textbf{Ground Truth Depth.} Given LiDAR data, intrinsics and extrinsics, we first compute the poses of LiDAR using the FAST-LIO algorithm \cite{xu2021fast}. Then, considering the sparsity of a single LiDAR point cloud, we utilize the poses to fuse the LiDAR point clouds within a certain period centered at current timestamp into a local dense cloud. Finally, we use calibration parameters of cameras and LiDAR to project the local fused point cloud into the camera coordinate system for ground truth depth. However, due to the sparsity of LiDAR point clouds, directly projecting them into the camera coordinate system may cause some non-visible points to appear. To this end, we propose to perform outlier removal with reference depth. Specifically, we introduce a foundation depth model, namely Metric3D \cite{yin2023metric3d, hu2024metric3d}, to estimate the reference depth for filtering the projected coarse depth. After the outlier removal, we obtain the final ground truth depth.           
            
            \textbf{Ground Truth Optical Flow.} In Fig. \ref{Fig:GT_Generation}, we further illustrate the generation of ground truth optical flow, which is calculated by ground truth depth with the poses and calibration parameters of the camera as follows:
            \begin{equation}
                \label{F}
                    F = KTD(X)K^{-1}X - X
            \end{equation}
            where $X$ is the pixel coordinate, $K$ is the camera intrinsics, $D$ is the depth map, and $T$ is the camera transformation, which is computed from the LiDAR poses and the extrinsics between the LiDAR and camera.
            
            \textbf{Event-Frame Fusion for Nighttime Scene.} Nighttime label generation is a challenging problem, since low light weakens the texture captured by the frame camera due to its low dynamic range. In contrast, event camera has the advantage of high dynamic range, motivating us to introduce an event-frame fusion for improving the accuracy of nighttime label generation. Specifically, in Fig. \ref{Fig:GT_Generation}, we take the low-light enhancement method \cite{liang2023coherent} to fuse the event and frame data for the visibility of the nighttime frame, and then estimate the reference depth from the enhanced image using Metric3D \cite{yin2023metric3d, hu2024metric3d}. As shown in Fig. \ref{Fig:Enhance_Compare}, the event-frame fusion can well enhance the low-light image, thus improving the details of the corresponding reference depth. Therefore, this fusion approach leverages the advantages of the coaxial devices, making it possible to generate accurate ground truth depth and flow under all-day conditions in Fig. \ref{Fig:GT_Visual}.
        
    
    \section{EXPERIMENTS}
        In this section, we conduct experiments to demonstrate that the proposed dataset can improve the performance and generalization of multimodal fusion. The results are shown in Table \ref{Tab:Experiments_Fusion} and \ref{Tab:Experiments_Night}, where F, E, and EF indicate training on the frame, event, and event-frame data respectively.

        \begin{table}
            \vspace{5pt}
            \caption
            {
                Discussion on effect of parallel and coaxial event-frame placement strategies on multimodal fusion.
            }
            \label{Tab:Experiments_Fusion}
            \centering
            \setlength\tabcolsep{3.5pt}
            \renewcommand{\arraystretch}{1.3}
            \begin{tabular}{cccc|ccc|ccc}
                \bottomrule
                    \multicolumn{2}{c}{\makecell{\textbf{Parallel} \\ \textbf{(DSEC)}}} & \multicolumn{2}{c|}{\makecell{\textbf{Coaxial} \\ \textbf{(CoSEC)}}} & 
                    \multicolumn{3}{c|}{\textbf{DSEC}} & \multicolumn{3}{c}{\textbf{CoSEC}} \\
                    Frame & Event & Frame & Event & 10m & 20m & 30m & 10m & 20m & 30m \\
                \hline
                    \faCheck & \usym{2613} & \faCheck & \usym{2613} & 2.56 & 2.82 & 3.86 & 2.57 & 2.79 & 5.61 \\
                    \faCheck & \faCheck & \faCheck & \usym{2613} & 1.18 & 1.56 & 2.81 & 1.75 & 2.00 & 5.15 \\
                    \faCheck & \faCheck & \faCheck & \faCheck & 1.14 & 1.52 & 2.79 & 1.68 & 1.92 & 5.01 \\
                    \faCheck & \usym{2613} & \faCheck & \faCheck & \textbf{1.11} & \textbf{1.46} & \textbf{2.65} & \textbf{1.66} & \textbf{1.90} & \textbf{4.94} \\
                \toprule
            \end{tabular}
        \end{table}
    
        \subsection{Experiment Setup}
            \textbf{Dataset.} We conduct experiments on DSEC \cite{Gehrig21ral} and CoSEC datasets. The event-frame data of the DSEC dataset is spatially unaligned due to the parallel strategy, while CoSEC dataset implements the coaxial strategy to capture aligned data. Note that the training and testing sets from CoSEC are selected to match the size of those from DSEC, and the comparisons should be performed within each dataset due to differences in scene scale caused by different capturing locations and camera intrinsics. 
            
            \textbf{Method.} To demonstrate the effectiveness of the pixel-level spatial alignment between different modalities on multimodal fusion, we take depth estimation as the example task, where we introduce the multimodal-based model RAMNet \cite{gehrig2021combining} to conduct experiments. We pre-train the depth model for weight initialization on dense depth maps from the EventScape \cite{gehrig2021combining} dataset, which is a synthetic dataset collected from the CARLA simulator \cite{dosovitskiy2017carla}. For the comparison experiment of different event-frame placement strategies in Table \ref{Tab:Experiments_Fusion}, we first train the frame branch of the RAMNet individually on the mixed frame data of DSEC and CoSEC. Then, we load the obtained weights of the frame branch into RAMNet, which is fine-tuned on the event-frame data of DSEC, CoSEC, and their mixture respectively. For the generalization experiment for nighttime scenes in Table \ref{Tab:Experiments_Night}, we train RAMNet and its unimodal-based baselines on nighttime data of DSEC and CoSEC.
            
            \textbf{Metric.} We choose the average absolute depth error for ground truth depth up to 10 m, 20 m and 30 m as quantitative metric \cite{gehrig2021combining}, which is lower for better performance.
            
        \subsection{Effectiveness of Coaxial Device on Multimodal Fusion}
            To validate the effectiveness of the coaxial device on event-frame fusion in Table \ref{Tab:Experiments_Fusion}, we compare unimodal-based and multimodal-based depth estimation models trained on DSEC \cite{Gehrig21ral} and CoSEC datasets. Note that the event-frame data of DSEC is captured via parallel strategy, and that of CoSEC is captured via coaxial strategy. First, multimodal fusion significantly improves the performance of depth estimation over the unimodal approach. This is because the complementary knowledge fusion between frame and event cameras can compensate for the limitation of frame-based imaging mechanism, thus improving the performance of depth estimation. Second, the depth models trained on the multimodal data of CoSEC perform better than those trained on the DSEC. The main reason is that the coaxial strategy used in CoSEC enables better pixel-level spatial alignment between the event and frame data, facilitating the multimodal-based depth estimation. Therefore, the proposed coaxial device ensures pixel-level alignment between event and frame, promoting the multimodal fusion for vision tasks.
            
        \subsection{Improvement of Generalization for Nighttime Scenes}
            To demonstrate the impact of the proposed dataset CoSEC on model generalization for nighttime scenes in Table \ref{Tab:Experiments_Night}, we evaluate the depth models trained on unimodal and multimodal nighttime data of DSEC \cite{Gehrig21ral} and CoSEC. First, the multimodal-based model trained on CoSEC outperforms the one trained on DSEC. The main reason is that the pixel-level spatially aligned event-frame data of CoSEC enhances multimodal fusion for nighttime scenes more effectively than the unaligned data of DSEC. Second, the models trained on nighttime data of CoSEC perform better than those trained on DSEC. This is because our nighttime enhancement strategy, namely event-frame fusion, can effectively improve the quality of ground truth depth in nighttime scenes. Therefore, the proposed CoSEC dataset can not only enhance the performance of multimodal fusion, but also facilitate the improvement of model generalization for nighttime scenes.

    \begin{table}
        \vspace{5pt}
        \caption
        {
            Discussion on improvement of the proposed dataset on the generalization of depth model for nighttime scenes.
        }
        \label{Tab:Experiments_Night}
        \centering
        \renewcommand{\arraystretch}{1.3}
        \begin{tabular}{l|ccc|ccc}
            \bottomrule
                \multicolumn{1}{c|}{\multirow{2} * {\textbf{Strategy}}} &
                \multicolumn{3}{c|}{\textbf{DSEC}} & \multicolumn{3}{c}{\textbf{CoSEC}} \\
                & 10m & 20m & 30m & 10m & 20m & 30m \\
            \hline
                \textbf{F\textsubscript{DSEC}} & 3.46 & 3.56 & 4.02 & 6.16 & 6.27 & 7.20 \\
                \textbf{F\textsubscript{CoSEC}} & \textbf{3.17} & \textbf{3.29} & \textbf{3.82} & \textbf{3.93} & \textbf{4.07} & \textbf{5.21} \\
            \hline
                \textbf{E\textsubscript{DSEC}} & 5.74 & 5.81 & 6.13 & 2.30 & 2.46 & 4.29 \\
                \textbf{E\textsubscript{CoSEC}} & \textbf{3.29} & \textbf{3.40} & \textbf{3.92} & \textbf{2.11} & \textbf{2.29} & \textbf{4.05} \\
            \hline
                \textbf{EF\textsubscript{DSEC}} & 4.34 & 4.45 & 4.85 & 1.47 & 1.65 & 3.59 \\
                \textbf{EF\textsubscript{CoSEC}} & \textbf{1.72} & \textbf{1.85} & \textbf{2.45} & \textbf{0.77} & \textbf{0.97} & \textbf{3.03} \\
            \toprule
        \end{tabular}
    \end{table}
    
    
    \section{CONCLUSION}
        In this work, we propose a coaxial stereo event camera dataset for autonomous driving, which is collected by the proposed multimodal system, including frame, event, LiDAR and INS. As for the system, we design the coaxial event-frame device to build the multimodal system, promoting the pixel-level spatial alignment between different sensors. As for the dataset, we introduce the aligned event-frame fusion approach to improve the accuracy of ground truth depth and optical flow generation in all-day scenes. Moreover, we conduct experiments to verify the superiority of the proposed multimodal dataset on improving the performance and generalization of multimodal fusion. In the future, we will extend the proposed dataset to more challenging scenes, such as adverse weather. We believe that the proposed dataset can facilitate the relevant research on multimodal fusion for the 3D dynamic scene perception.

    
    \section*{ACKNOWLEDGMENT}
        This work was supported in part by the National Natural Science Foundation of China under Grant 62371203. The computation is completed in the HPC Platform of Huazhong University of Science and Technology.

    
    \addtolength{\textheight}{-6cm}
    \bibliographystyle{IEEEtranBST/IEEEtran}
    \bibliography{references}
\end{document}